\documentclass[11pt,a4paper]{article}
\usepackage[hyperref]{acl2020}
\usepackage{times}
\usepackage{latexsym}
\usepackage{boldline}

\usepackage{amsmath,amsfonts,bm}

\def\eqref#1{equation~\ref{#1}}

\def\1{\bm{1}}

\DeclareMathAlphabet{\mathsfit}{\encodingdefault}{\sfdefault}{m}{sl}
\SetMathAlphabet{\mathsfit}{bold}{\encodingdefault}{\sfdefault}{bx}{n}

\usepackage{hyperref}
\usepackage{url}
\usepackage{amsmath}
\usepackage{multirow}
\usepackage{multicol}
\usepackage{CJKutf8}
\usepackage{subfigure}
\usepackage{graphicx}
\usepackage{makecell}
\usepackage{tabularx}
\usepackage{float}
\usepackage{latexsym}
\usepackage{booktabs}
\usepackage{tablefootnote}
\usepackage{amsmath}
\usepackage{lscape}
\usepackage{multirow}
\usepackage{amsmath}

\usepackage{mathtools}

\aclfinalcopy 

\title{Character-Level Translation with Self-attention}

\author{Yingqiang Gao$^{\dagger\ddagger}$, Nikola I. Nikolov$^\ddagger$, Yuhuang Hu$^\ddagger$, Richard H.R. Hahnloser$^\ddagger$ \\
  $^\dagger$Department of Informatics, Technical University of Munich \\ $^\ddagger$Institute of Neuroinformatics,
  University of Zurich and ETH Zurich \\
  \texttt{yingqiang.gao@in.tum.de} \hspace{0.5cm}\texttt{\{niniko, yuhuang.hu, rich\}@ini.ethz.ch} } 

\date{}

\begin{document}

\maketitle

\begin{abstract}

We explore the suitability of self-attention models for character-level neural machine translation. We test the standard transformer model, as well as a novel variant in which the encoder block combines information from nearby characters using convolutions. We perform extensive experiments on WMT and UN datasets, testing both bilingual and multilingual translation to English using up to three input languages (French, Spanish, and Chinese). Our transformer variant consistently outperforms the standard transformer at the character-level and converges faster while learning more robust character-level alignments.\footnote{Code available at \url{https://github.com/CharizardAcademy/convtransformer}} 

\end{abstract}

\section{Introduction}

Most existing Neural Machine Translation (NMT) models operate on the word or subword-level, which tends to make these models memory inefficient because of large vocabulary sizes. Character-level models \cite{lee2017fully,cherry2018revisiting} instead work directly on raw characters, resulting in a more compact language representation, while mitigating out-of-vocabulary (OOV) problems~\cite{luong-manning-2016-achieving}. Character-level models are also very suitable for multilingual translation since multiple languages can be modeled using the same character vocabulary. Multilingual training can lead to improvements in the overall performance without an increase in model complexity \cite{lee2017fully}, while also circumventing the need to train separate models for each language pair. 

Models based on self-attention have achieved excellent performance on a number of tasks, including machine translation \cite{vaswani2017attention} and representation learning \cite{devlin2018bert,yang2019xlnet}. Despite the success of these models, their suitability for character-level translation remains largely unexplored, with most efforts having focused on recurrent models (\emph{e.g.}, \citet{lee2017fully,cherry2018revisiting}). 

\begin{figure}
    \centering
    \includegraphics[width=\linewidth]{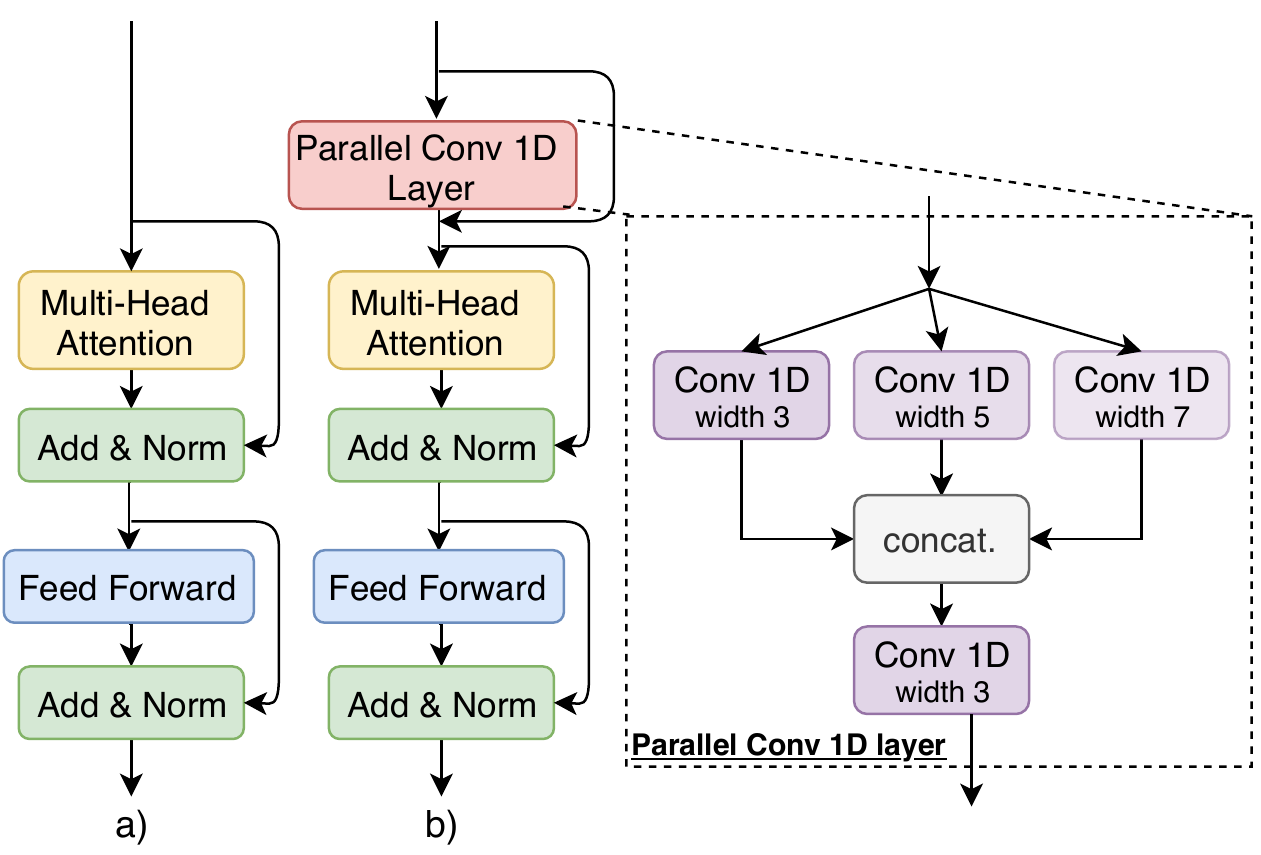}
    \caption{A comparison of the encoder blocks in the standard transformer (a) and our novel modification, the convtransformer (b), which uses 1D convolutions to facilitate character interactions.}
    \label{fig:archi}
\end{figure}

In this work, we perform an in-depth investigation of the suitability of self-attention models for character-level translation. We consider two models: the standard transformer from \citet{vaswani2017attention} and a novel variant that we call the \textit{convtransformer} (Figure \ref{fig:archi}, Section \ref{sec:convtransformer}). The convtransformer uses convolutions to facilitate interactions among nearby character representations.

We evaluate these models on both bilingual and multilingual translation to English, using up to three input languages: French (FR), Spanish (ES), and Chinese (ZH). We compare the performance when translating from close (\emph{e.g.}, FR and ES) and on distant (\emph{e.g.}, FR and ZH) input languages (Section~\ref{subsec:aumatic:evaluation}) and we analyze the learned character alignments (Section~\ref{subsec:learned:alignments}). We find that self-attention models work surprisingly well for character-level translation, achieving competitive performance to equivalent subword-level models while requiring up to 60\% fewer parameters (under the same model configuration). At the character-level, the convtransformer outperforms the standard transformer, converges faster, and produces more robust alignments.

\section{Background}

\subsection{Character-level NMT}

Fully character-level translation was first tackled in \citet{lee2017fully}, who proposed a recurrent encoder-decoder model. Their encoder combines convolutional layers with max-pooling and highway layers to construct intermediate representations of segments of nearby characters. Their decoder network autoregressively generates the output translation one character at a time, utilizing attention on the encoded representations. 

\citet{lee2017fully}'s approach showed promising results on \emph{multilingual translation} in particular. Without any architectural modifications or changes to the character vocabularies, training on multiple source languages yielded performance improvements while also acting as a regularizer. Multilingual training of character-level models is possible not only for languages that have almost identical character vocabularies, such as French and Spanish, but even for distant languages that can be mapped to a common character-level vocabulary, for example, through latinizing Russian \cite{lee2017fully} or Chinese \cite{nikolov2018character}.

More recently, \cite{cherry2018revisiting} performed an in-depth comparison between different character- and subword-level models. They showed that, given sufficient computational time and model capacity, character-level models can outperform subword-level models, due to their greater flexibility in processing and segmenting the input and output sequences. 

\subsection{The Transformer}

The transformer \cite{vaswani2017attention} is an attention-driven encoder-decoder model that has achieved state-of-the-art performance on a number of sequence modeling tasks in NLP. Instead of using recurrence, the transformer uses only feedforward layers based on self-attention. The standard transformer architecture consists of six stacked encoder layers that process the input using self-attention and six decoder layers that autoregressively generate the output sequence.

The original transformer \cite{vaswani2017attention} computes a scaled dot-product attention by taking as input query $Q$, key $K$, and value $V$ matrices: 
\begin{equation*}
    \text{Attention}(Q,K,V) = \text{softmax}\left(\frac{QK^T}{\sqrt{d_k}} \right)V,
\end{equation*}
where $\sqrt{d_k}$ is a scaling factor. For the encoder, $Q$, $K$ and $V$ are equivalent, thus, given an input sequence with length $N$, $\text{Attention}$ performs $N^2$ comparisons, relating each word position with the rest of the words in the input sequence. In practice, $Q$, $K$, and $V$ are projected into different representation subspaces (called heads), to perform Multi-Head Attention, with each head learning different word relations, some of which might be interpretable \cite{vaswani2017attention,voita-etal-2019-analyzing}. 

Intuitively, attention as an operation might not be as meaningful for encoding individual characters as it is for words, because individual character representations might provide limited semantic information for learning meaningful relations on the sentence level. However, recent work on language modeling \cite{al2019character} has surprisingly shown that attention can be very effective for modeling characters, raising the question of how well the transformer would work on character-level bilingual and multilingual translation, and what architectures would be suitable for this task. These are the questions this paper sets out to investigate.

\section{Convolutional Transformer}\label{sec:convtransformer}

To facilitate character-level interactions in the transformer, we propose a modification of the standard architecture, which we call the \textit{convtransformer}. In this architecture, we use the same decoder as the standard transformer, but we adapt each encoder block to include an additional sub-block. The sub-block (Figure \ref{fig:archi}, b), inspired from \citet{lee2017fully}, is applied to the input representations $M$, before applying self-attention. The sub-block consists of three 1D convolutional layers, $C_w$, with different context window sizes $w$. In order to maintain the temporal resolution of convolutions, the padding is set to $\lfloor \frac{w - 1}{2} \rfloor$. 

We apply three separate convolutional layers, $C_3$, $C_5$ and $C_7$, in parallel, using context window sizes of $3$, $5$ and $7$, respectively. The different context window sizes aim to resemble character-level interactions of different levels of granularity, such as on the subword- or word-level. To compute the final output of the convolutional sub-block, the outputs of the three layers are concatenated and passed through an additional 1D convolutional layer with context window size $3$, $C_3^{'}$, which fuses the representations:
\begin{align*}
    &\text{Conv}(M) = \\
    & M + C_3^{'} (\text{Concat}(C_3(M), C_5(M), C_7(M))).
\end{align*}

For all convolutional layers, we set the number of filters to be equal to the embedding dimension size $d_{\text{model}}$, which results in an output of equal dimension as the input $M$. Therefore, in contrast to \citet{lee2017fully}, who use max-pooling to compress the input character sequence into segments of characters, here we leave the resolution unchanged, for both transformer and convtransformer models. Finally, for additional flexibility, we add a residual connection \citep{he2016deep} from the input to the output of the convolutional block.

\section{Experimental Set-up}

\paragraph{Datasets.}

We conduct experiments on two datasets. First, we use the \textbf{WMT15 DE$\rightarrow$EN} dataset, on which we test different model configurations and compare our results to previous work on character-level translation. We follow the preprocessing in \citet{lee2017fully} and use the newstest-2014 dataset for testing. Second, we conduct our main experiments using the \textbf{United Nations Parallel Corporus} (UN) \cite{ziemski2016united}, for two reasons: (i) UN contains a large number of parallel sentences from six languages, allowing us to conduct multilingual experiments; (ii) all sentences in the corpus are from the same domain. We construct our training corpora by randomly sampling one million sentence pairs from the FR, ES, and ZH parts of the UN dataset, targeting translation to English. To construct multilingual datasets, we combine the respective bilingual datasets (\emph{e.g.}, FR$\rightarrow$EN, and ES$\rightarrow$EN) and shuffle them. To ensure all languages share the same character vocabulary, we latinize the Chinese dataset using the Wubi encoding method, following \cite{nikolov2018character}. For testing, we use the original UN test sets provided for each pair. 

\paragraph{Tasks.}

Our experiments are designed as follows: (i) bilingual scenario, in which we train a model with a single input language; (ii) multilingual scenario, in which we input two or three languages at the same time without providing any language identifiers to the models and without increasing the number of parameters. We test combining input languages that can be considered as more similar in terms of syntax and vocabulary (\emph{e.g.} FR and ES) as well as more distant (\emph{e.g.}, ES and ZH). 

\section{Results} 

\renewcommand{\tabcolsep}{3pt}

\begin{table}
    \centering
    \small 
    \begin{tabular}{c|lcc|}
        \clineB{2-4}{1} & \textbf{Model} & \textbf{BLEU} & \textbf{\#par} \\
         \hline
         \parbox[t]{2mm}{\multirow{5}{*}{\rotatebox[origin=c]{90}{\tiny{\textit{character-level}}}}} 
         & \citet{lee2017fully} & 25.77 & 69M  \\
         & transformer-6-layer & 28.8 & 49M \\
         &  convtransformer-6-layer & 29.23 & 68M \\
         & transformer-12-layer & 29.81 & 93M \\
         &  convtransformer-12-layer & 30.16 & 131M \\
        \hline
       \parbox[t]{2mm}{\multirow{2}{*}{\rotatebox[origin=c]{90}{\tiny{\textit{bpe}}}}} 
        & transformer-6-layer & 30.06 & 121M \\ 
        & transformer-12-layer & 31.60 & 165M \\ \hline
    \end{tabular}
    \caption{Comparison of architecture variants on the WMT15 DE$\rightarrow$EN dataset. \textbf{\#par} is the number of model parameters.}
    \label{tab:wmt_results}
\end{table}

\subsection{Automatic evaluation}\label{subsec:aumatic:evaluation}

\paragraph{Model comparison.} In Table \ref{tab:wmt_results}, we compare the BLEU performance \cite{papineni2002bleu} of diverse character-level architectures trained on the WMT dataset. For reference, we include the recurrent character-level model from \citet{lee2017fully}, as well as transformers trained on the subword level using a vocabulary of 50k byte-pair encoding (BPE) tokens \cite{sennrich2015neural}. All models were trained on four Nvidia GTX 1080X GPUs for 20 epochs.

We find character-level training to be 3 to 5 times slower than subword-level training due to much longer sequence lengths. However, the standard transformer trained at the character level already achieves very good performance, outperforming the recurrent model from \citet{lee2017fully}. On this dataset, our convtransformer variant performs on par with the character-level transformer. Character-level transformers also perform competitively with equivalent BPE models while requiring up to 60\% fewer parameters. Furthermore, our 12-layer convtransformer model matches the performance of the 6-layer BPE transformer, which has a comparable number of parameters. 

\paragraph{Multilingual experiments.} In Table \ref{tab:un_results}, we report our BLEU results on the UN dataset using the 6-layer transformer/convtransformer character-level models (Appendix \ref{sec:examples} contains example model outputs). All of our models were trained for 30 epochs. Multilingual models are evaluated on translation from all possible input languages to English.

\renewcommand{\tabcolsep}{2pt}

\begin{table}
\centering
\small
\begin{tabular}{c|c|c|ccc|ccc|}
\clineB{2-9}{1} 
& \textbf{Model} &
\textbf{\#P} & \multicolumn{3}{c|}{\textbf{transformer}} & \multicolumn{3}{c|}{\textbf{convtransformer}} \\ 
 & \textbf{Input lang.} & & \textit{t-FR} & \textit{t-ES} & \textit{t-ZH} & \textit{t-FR} & \textit{t-ES} & \textit{t-ZH} \\ \hline
\parbox[t]{2mm}{\multirow{3}{*}{\rotatebox[origin=c]{90}{\tiny{\textit{bilingual}}}}} & 
\textit{FR} & 1M & 32.48 & - & - & 33.69 & - & - \\ 
 & \textit{ES} & 1M & - & 39.90 & - & - & 41.41 & - \\ 
 & \textit{ZH} & 1M & - & - & 38.70 & - & - & \textbf{41.01} \\ \hline
\parbox[t]{2mm}{\multirow{4}{*}{\rotatebox[origin=c]{90}{\tiny{\textit{multilingual}}}}} & \textit{FR+ES} & 2M & 33.51 & 40.83 & - & \textbf{34.69} & \textbf{41.84} & - \\ 
 & \textit{FR+ZH} & 2M & 32.89 & - & 37.92 & 33.98 & - & 40.56 \\ 
 & \textit{ES+ZH} & 2M & - & 40.43 & 38.23 & - & 41.49 & 40.41 \\ 
 \clineB{2-9}{1} 
 & \textit{FR+ES+ZH} & 3M & 33.69 & 40.71 & 38.01 & 34.38 & 41.73 & 39.87 \\ \hline
\end{tabular}
\caption{BLEU scores on the UN dataset, for different input training languages (first column), and evaluated on three different test sets (t-FR, t-ES and t-ZH). The target language is always English. \textbf{\#P} is the number of training pairs. The best overall results for each language are in bold.}
\label{tab:un_results}
\end{table}

Although multilingual translation can be realized using subword-level models through extracting a joint segmentation for all input languages (e.g., as in \citet{firat-etal-2016-multi,johnson2017google}), here we do not include any subword-level multilingual baselines, for two reasons. First, extracting a good multilingual segmentation is much more challenging for our choice of input languages, which includes distant languages such as Chinese and Spanish. Second, as discussed previously, subword-level models have a much larger number of parameters, making a balanced comparison with character-level models difficult. 

The convtransformer consistently outperforms the character-level transformer on this dataset, with a gap of up to 2.3 BLEU on bilingual translation (ZH$\rightarrow$EN) and up to 2.6 BLEU on multilingual translation (FR+ZH$\rightarrow$EN). Training multilingual models on similar input languages (FR + ES$\rightarrow$EN) leads to improved performance for both languages, which is consistent with \citep{lee2017fully}. Training on distant languages is surprisingly still effective in some cases. For example, the models trained on FR+ZH$\rightarrow$EN outperform the models trained just on FR$\rightarrow$EN; however they perform worse than the bilingual models trained on ZH$\rightarrow$EN. Thus, distant-language training seems to be helpful mainly when the input language is closer to the target translation language (which is English here). 

\begin{figure}
    \centering
    \includegraphics[width=\linewidth]{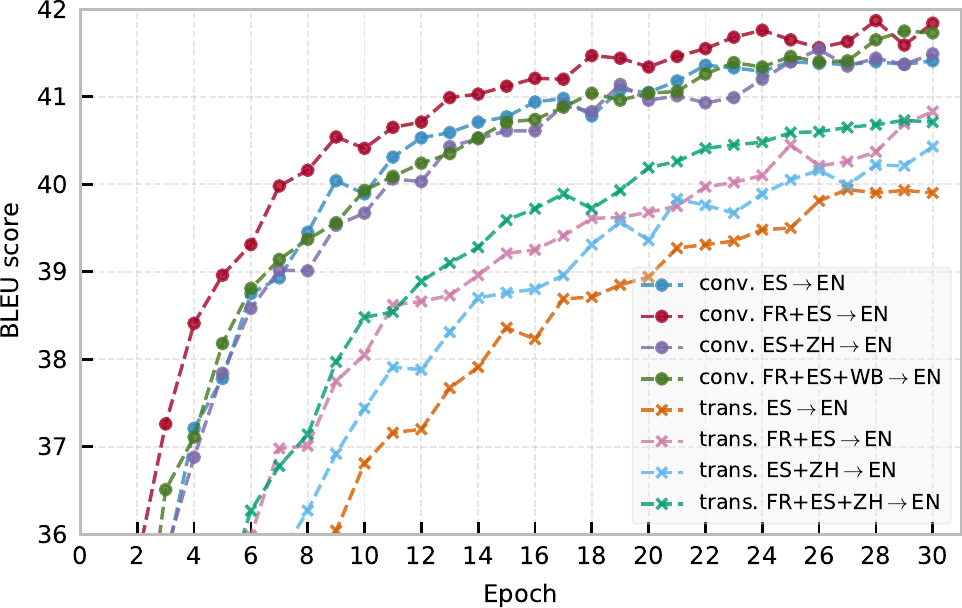}
    \caption{BLEU scores on the UN dataset as a function of epoch number, for bilingual and multilingual character-level translation from ES to EN. \texttt{conv.} is the convtransformer, while \texttt{trans.} is the original transformer.}
    \label{fig:convergence}
\end{figure}

The convtransformer is about 30\% slower to train than the transformer (see Figure \ref{fig:convergence}). Nevertheless, the convtransformer reaches comparable performance in less than half of the number of epochs, leading to an overall training speedup compared to the transformer. 

\begin{figure*}
    \centering
    \includegraphics[scale=0.6]{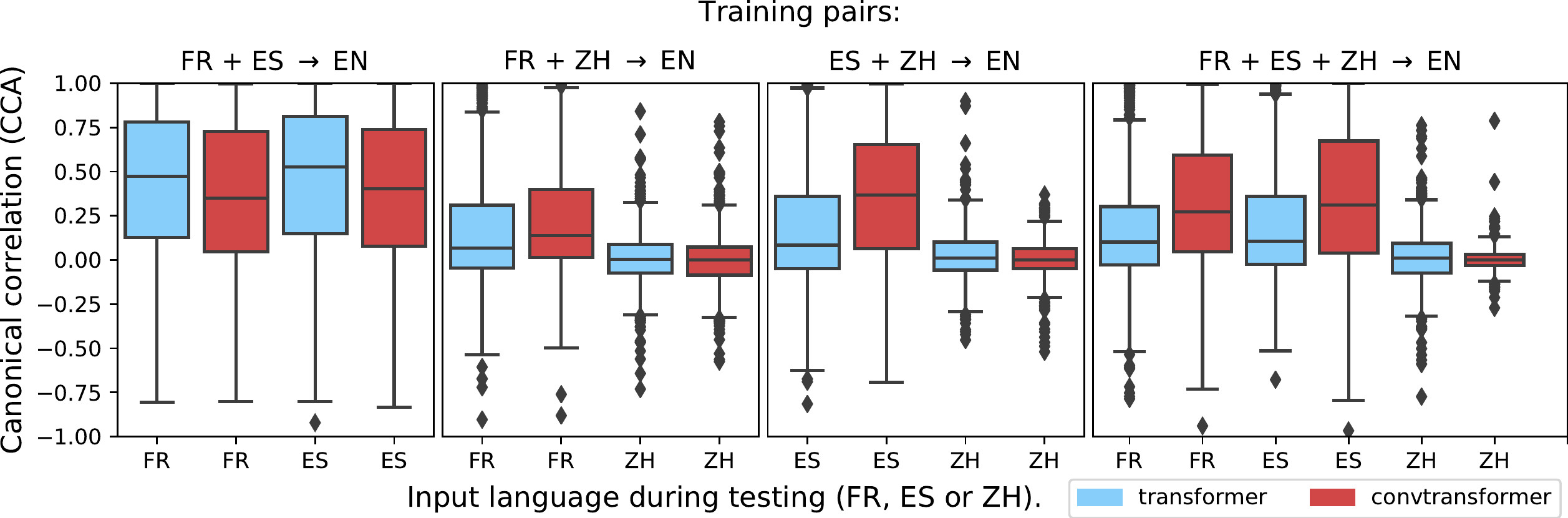}
    \caption{Canonical correlation between multilingual and bilingual translation models trained on the UN dataset.}
    \label{fig:cca}
\end{figure*}

\subsection{Analysis of Learned Alignments}\label{subsec:learned:alignments}

To gain a better understanding of the multilingual models, we analyze their learned character alignments as inferred from the model attention probabilities. For each input language (\emph{e.g.}, FR), we compare the alignments learned by each of our multilingual models (\emph{e.g.}, FR + ES $\rightarrow$ EN model) to the alignments learned by the corresponding bilingual model (\emph{e.g.}, FR $\rightarrow$ EN). Our intuition is that the bilingual models have the greatest flexibility to learn high-quality alignments because they are not distracted by other input languages. Multilingual models, by contrast,  might learn lower quality alignments because either (i) the architecture is not robust enough for multilingual training; or (ii) the languages are too dissimilar to allow for effective joint training, prompting the model to learn alternative alignment strategies to accommodate for all languages. 

We quantify the alignments using canonical correlation analysis (CCA) \cite{morcos2018insights}. First, we sample 500 random sentences from each of our UN testing datasets (FR, ES, or ZH) and then produce alignment matrices by extracting the encoder-decoder attention from the last layer of each model. We use CCA to project each alignment matrix to a common vector space and infer the correlation. We analyze our transformer and convtransformer models separately. Our results are in Figure \ref{fig:cca}, while Appendix \ref{sec:visualization} contains example alignment visualizations. 

For similar source and target languages (\emph{e.g.}, the FR+ES$\rightarrow$EN model), we observe a strong positive correlation to the bilingual models, indicating that alignments can be simultaneously learned. When introducing a distant source language (ZH) in the training, we observe a drop in correlation, for FR and ES, and an even larger drop for ZH. This result is in line with our BLEU results from Section \ref{subsec:aumatic:evaluation}, suggesting that multilingual training on distant input languages is more challenging than multilingual training on similar input languages. The convtransformer is more robust to the introduction of a distant language than the transformer ($p<0.005$ for FR and ES inputs, according to a one-way ANOVA test). Our results also suggest that more sophisticated attention architectures might need to be developed when training multilingual models on several distant input languages. 

\section{Conclusion}

We performed a detailed investigation of the utility of self-attention models for character-level translation. We test the standard transformer architecture, as well as introduce a novel variant which augments the transformer encoder with convolutions, to facilitate information propagation across nearby characters. Our experiments show that self-attention performs very well on character-level translation, with character-level architectures performing competitively when compared to equivalent subword-level architectures while requiring fewer parameters. Training on multiple input languages is also effective and leads to improvements across all languages when the source and target languages are similar. When the languages are different, we observe a drop in performance, in particular for the distant language. 

In future work, we will extend our analysis to include additional source and target languages from different language families, such as more Asian languages. We will also work towards improving the training efficiency of character-level models, which is one of their main bottlenecks, as well as towards improving their effectiveness in multilingual training. 

\section*{Acknowledgements}

We acknowledge support from the Swiss National Science Foundation (grant 31003A\_156976) and the National Centre of Competence in Research (NCCR) Robotics. We also thank the anonymous reviewers for their useful comments.

\bibliographystyle{bibliography}


\appendix
\section{Example model outputs}\label{sec:examples}

Tables \ref{tab:FR-EN}, \ref{tab:ES-EN}, and \ref{tab:CN-EN} contain example translations produced by our different bilingual and multilingual models trained on the UN datasets.  

\section{Visualization of Attention}\label{sec:visualization}

\begin{figure*}[!ht]
   \centering
   \subfigure[transformer trained on FR$\rightarrow$EN, testing with FR as input.]
    {
        \includegraphics[scale=.7]{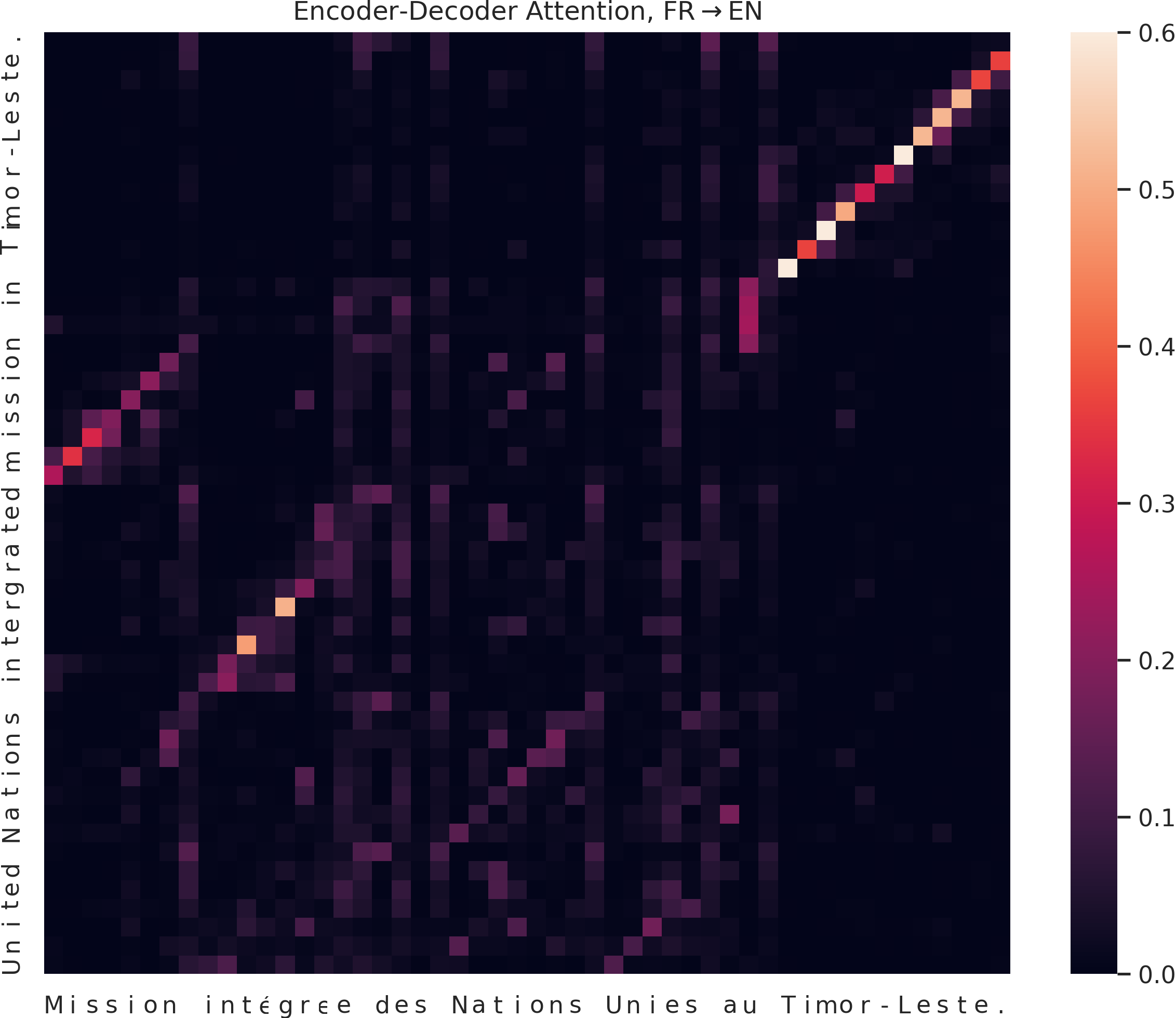}
    }
    \quad
    \subfigure[convtransformer trained on FR$\rightarrow$EN, testing with FR as input.]
    {
        \includegraphics[scale=.7]{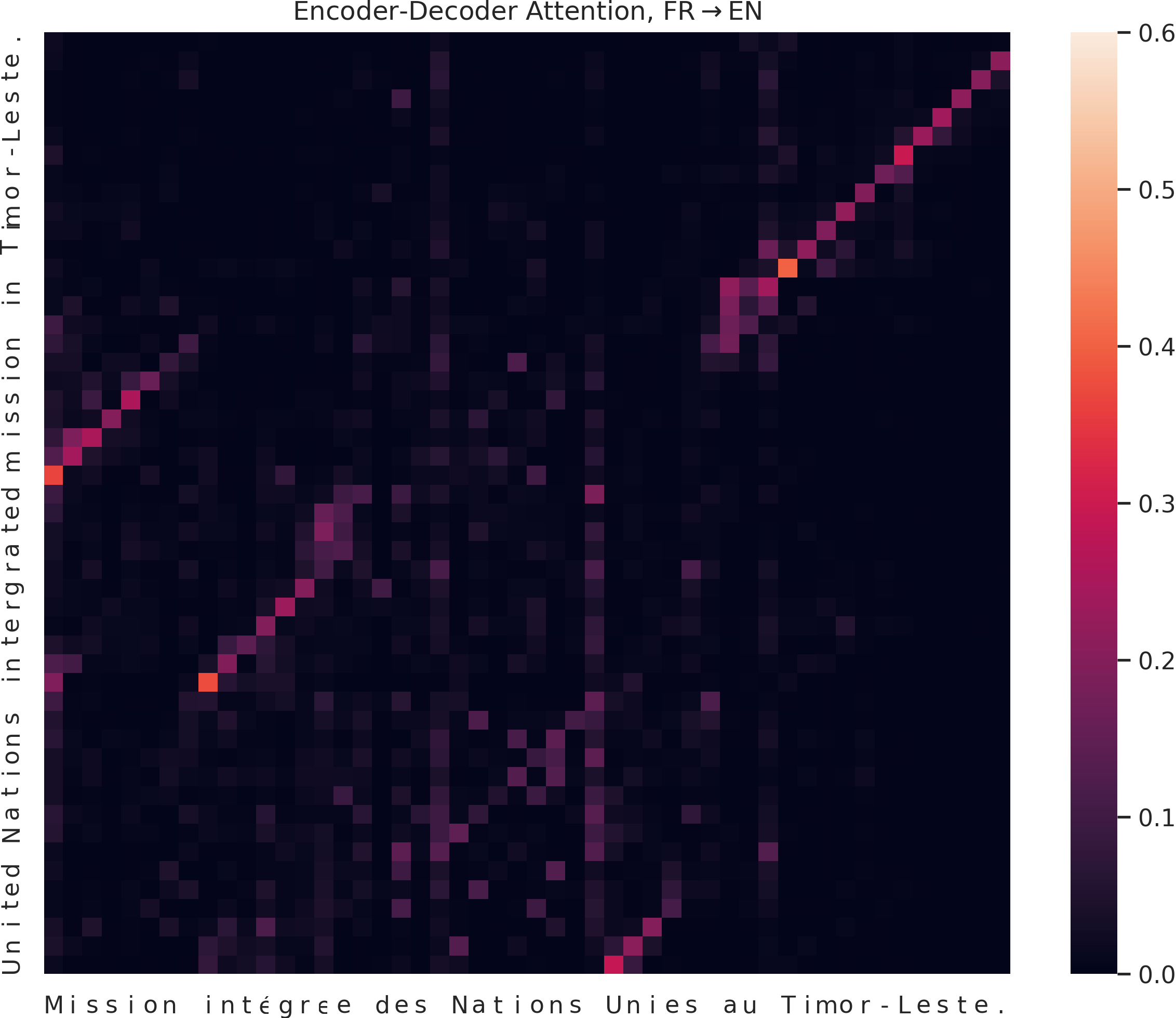}
    }
    \caption{
        Example alignments produced by character-level models trained on FR$\rightarrow$EN.
    }
    \label{fig:att-fr-en}
\end{figure*}

\begin{figure*}[!ht]
    \centering
    \subfigure[transformer trained on FR+ES$\rightarrow$EN, testing with FR as input.]
    {
        \includegraphics[scale=0.7]{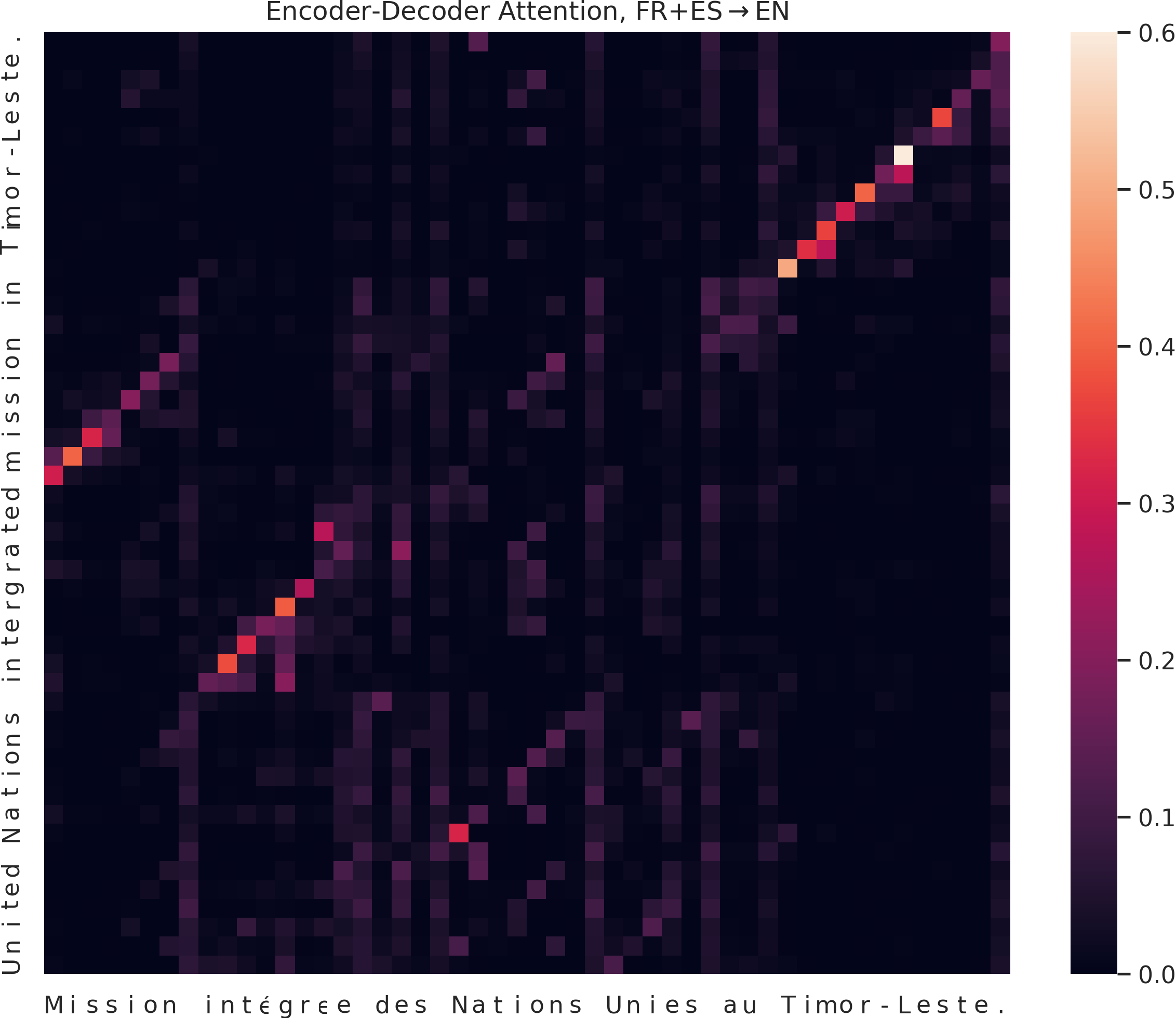}
    }
    \quad
    \subfigure[convtransformer trained on FR+ES$\rightarrow$EN, testing with FR as input]
    {
        \includegraphics[scale=0.7]{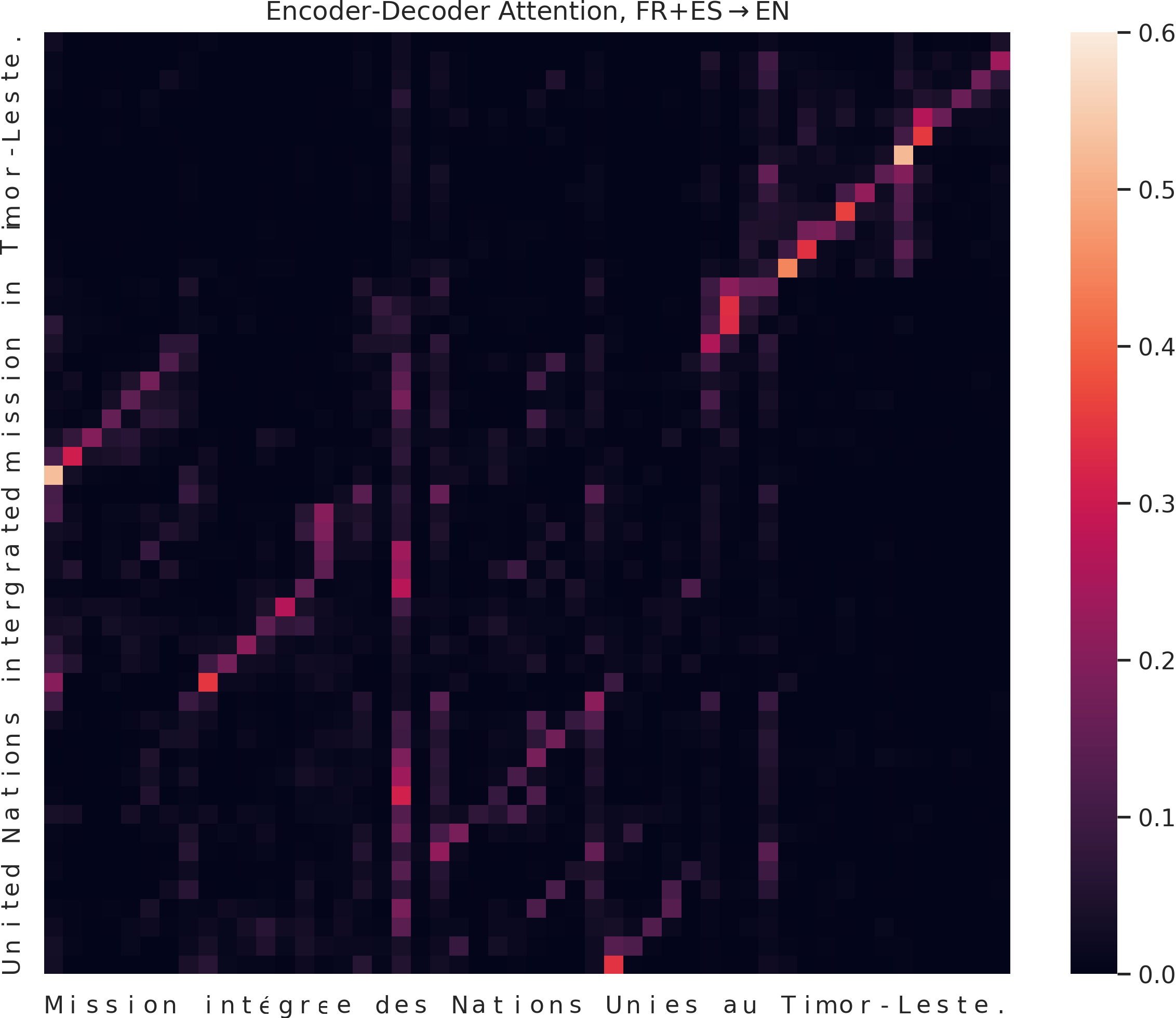}
    }
    \caption{
        Example alignments produced by character-level models trained on FR+ES$\rightarrow$EN. 
    }
    \label{fig:att-fres-en}
\end{figure*}

\begin{figure*}[!ht]
    \centering
    \subfigure[transformer FR+ZH$\rightarrow$EN, test on FR]
    {
        \includegraphics[scale=0.7]{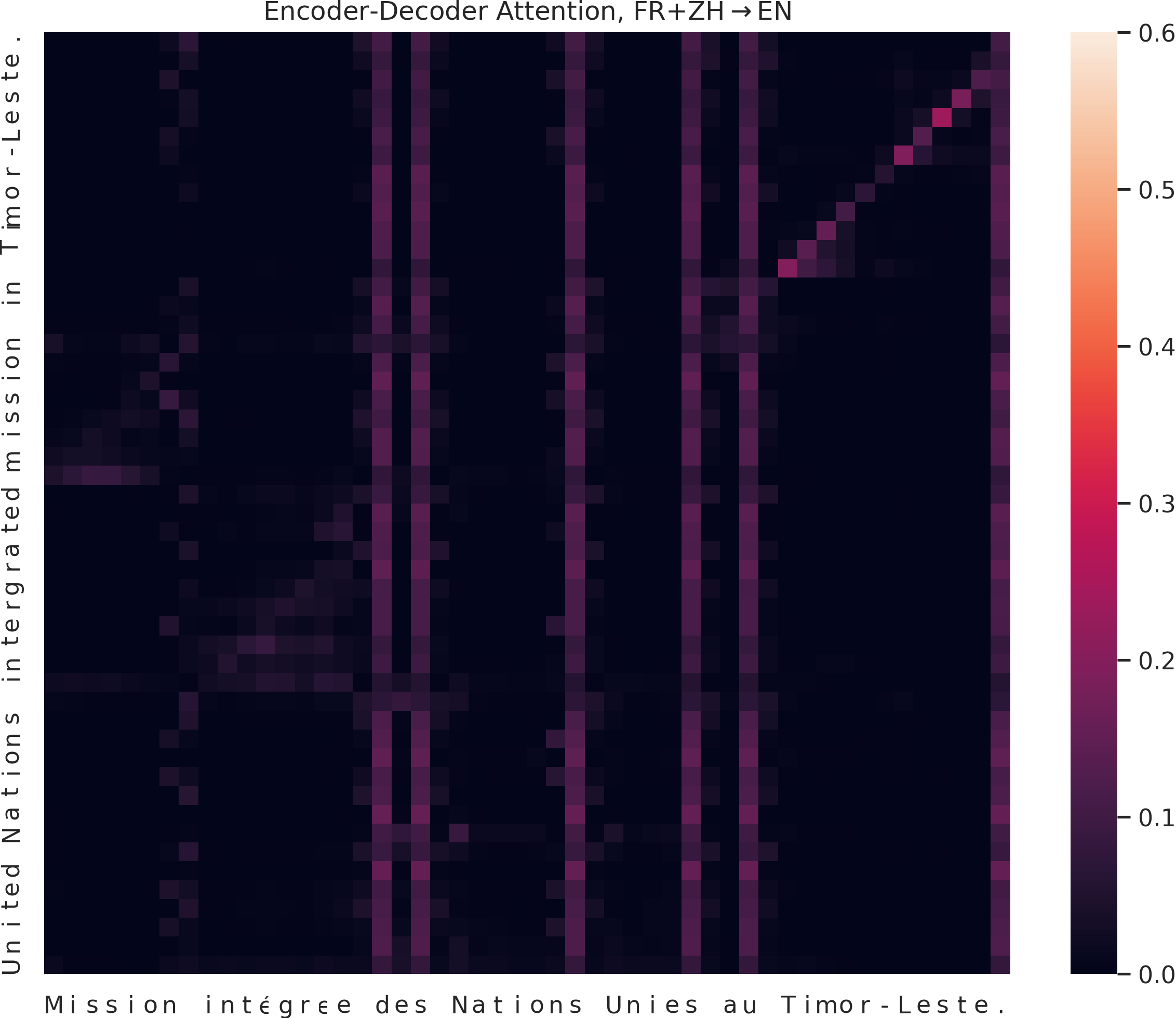}
    }
    \quad
    \subfigure[convtransformer FR+ZH$\rightarrow$EN, test on FR]
    {
        \includegraphics[scale=0.7]{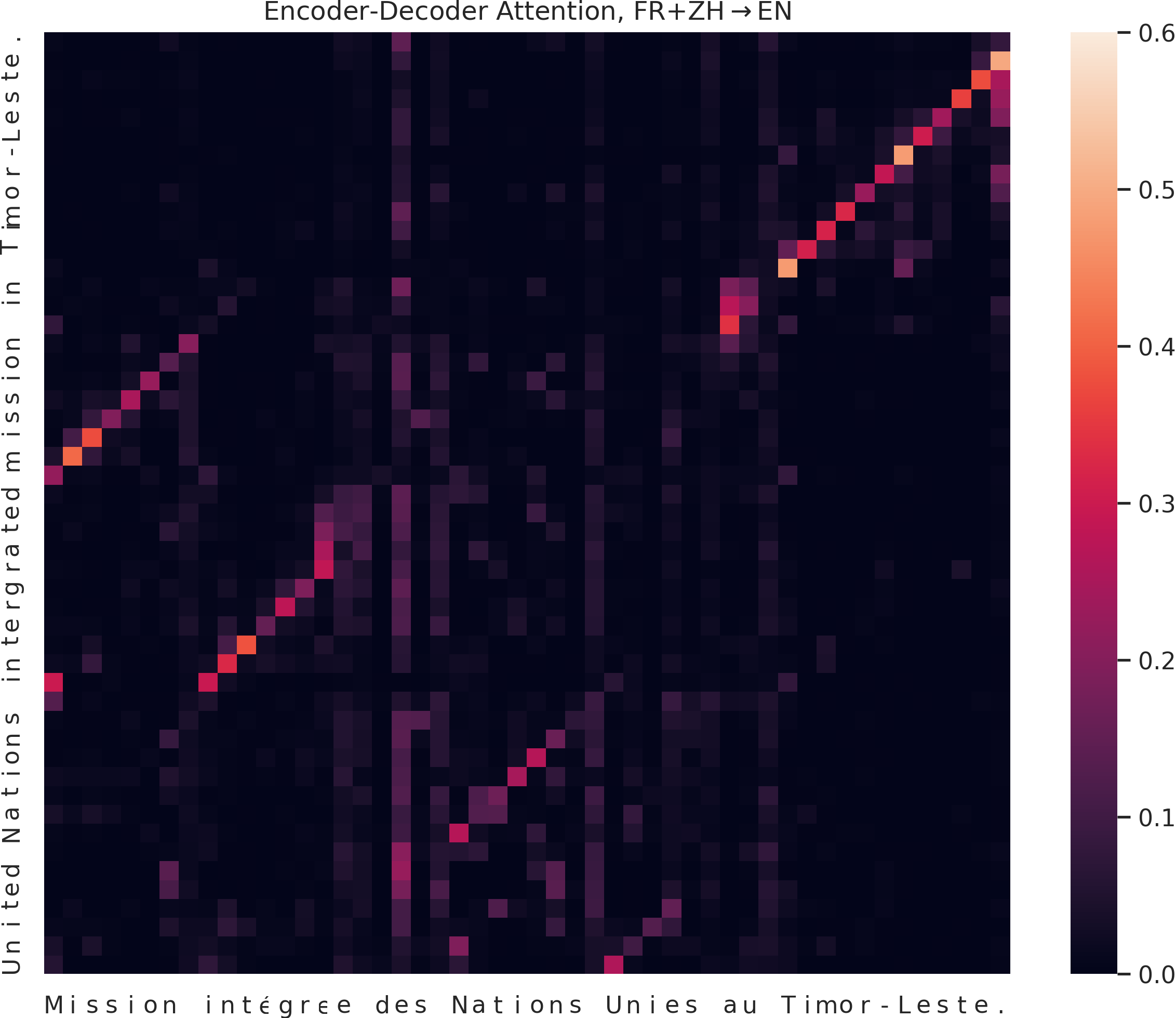}
    }
    \caption{
        Example alignments produced by character-level models trained on FR+ZH$\rightarrow$EN. 
    }
    \label{fig:att-frzh-en}
\end{figure*}

\begin{figure*}[!ht]
    \centering
    \subfigure[transformer FR+ES+ZH$\rightarrow$EN, test on FR]
    {
        \includegraphics[scale=0.7]{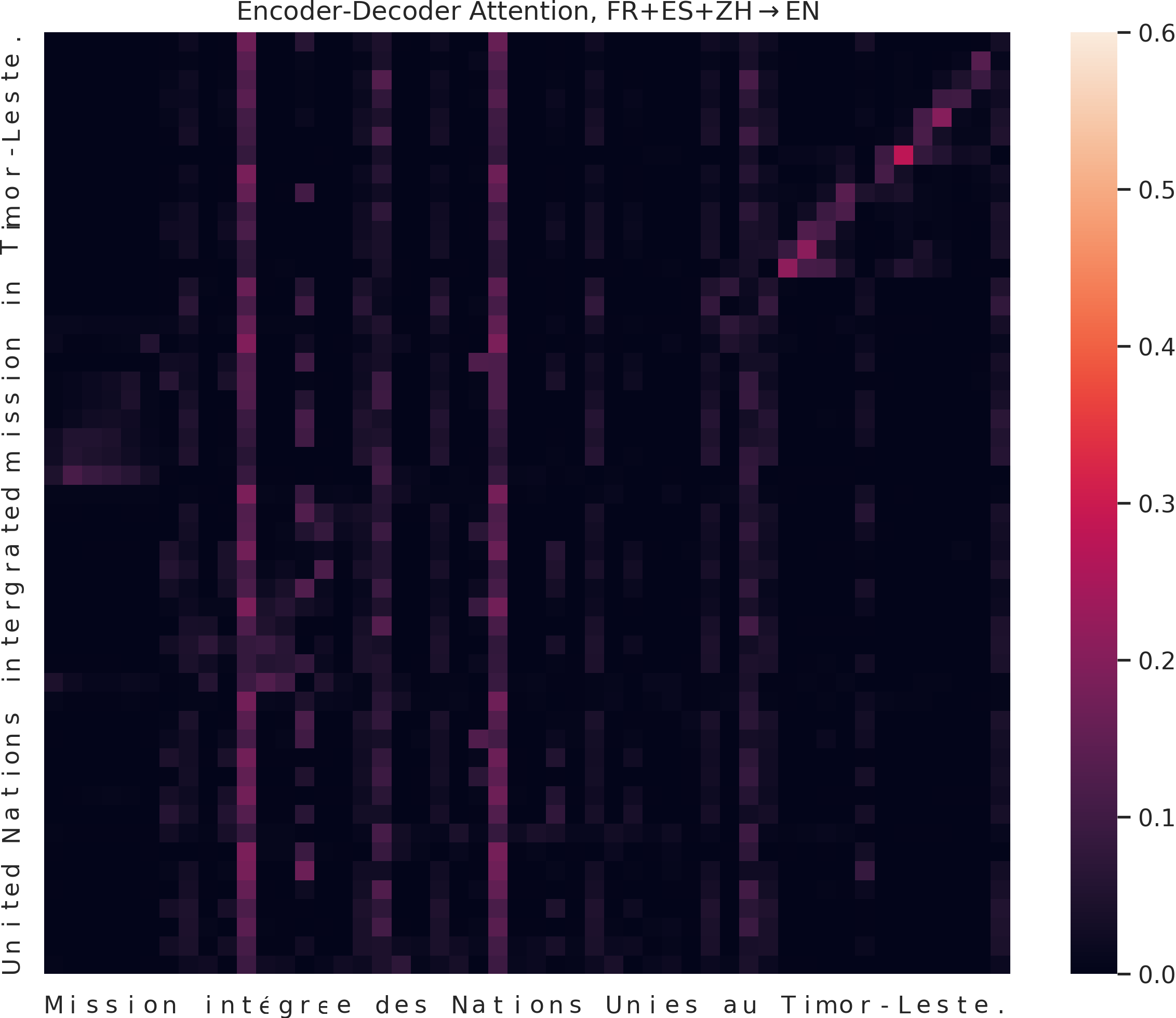}
    }
    \quad
    \subfigure[convtransformer FR+ES+ZH$\rightarrow$EN, test on FR]
    {
        \includegraphics[scale=0.7]{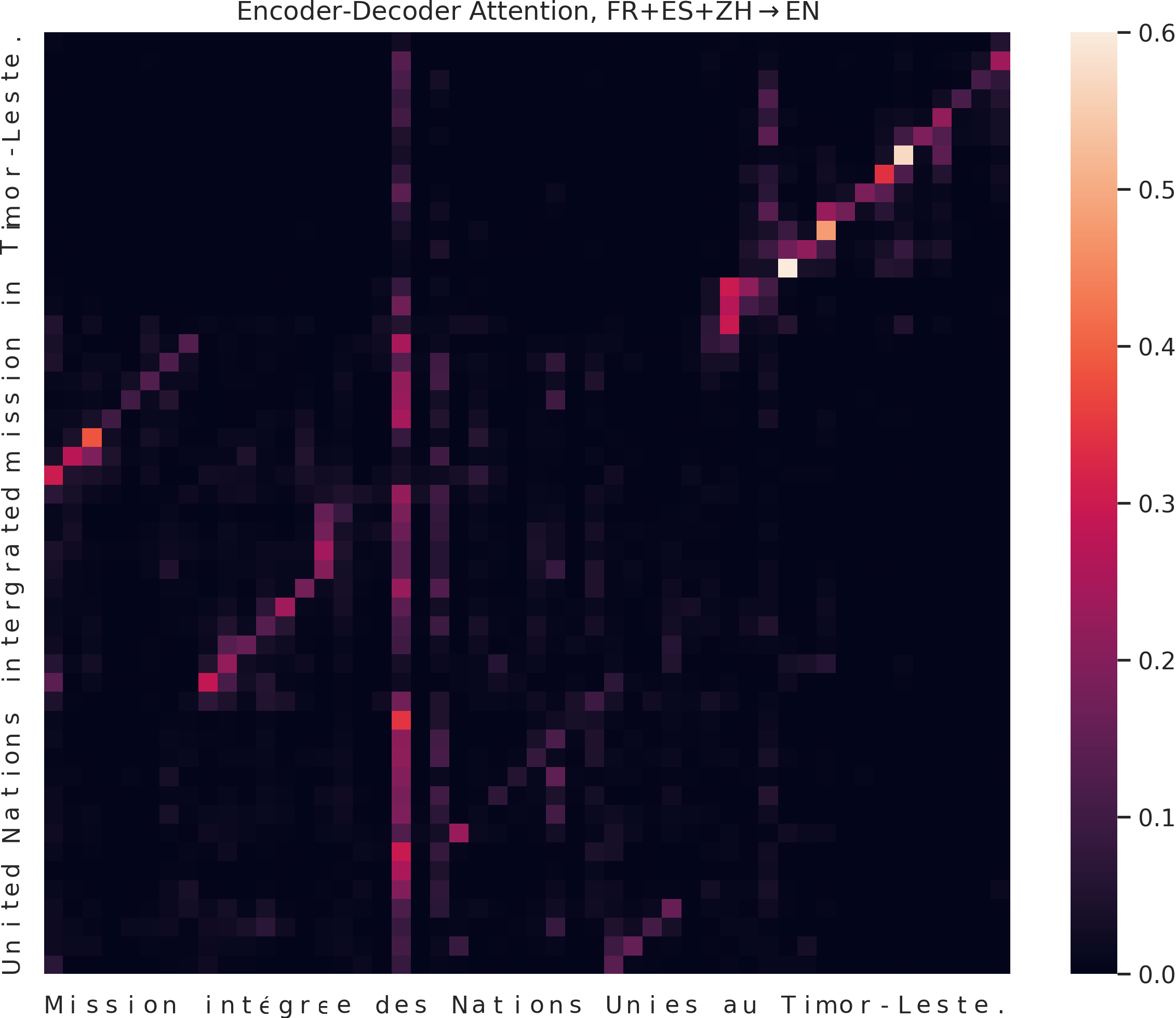}
    }
    \caption{
        Example alignments produced by character-level models trained on FR+ES+ZH$\rightarrow$EN. 
    }
    \label{fig:att-freszh-en}
\end{figure*}

In Figures \ref{fig:att-fr-en},\ref{fig:att-fres-en}, \ref{fig:att-frzh-en} and \ref{fig:att-freszh-en} we plot example alignments produced by our different bilingual and multilingual models trained on the UN datasets, always testing on translation from FR to EN. The alignments are produced by extracting the encoder-decoder attention of the last decoder layer of our transformer/convtransformer models. 

We observe the following patterns: ($i$) for bilingual translation (Figure \ref{fig:att-fr-en}), the convtransformer has a sharper weight distribution on the matching characters and words than the transformer; ($ii$) for multilingual translation of close languages (FR+ES$\rightarrow$EN, Figure \ref{fig:att-fres-en}), both transformer and convtransformer are able to preserve the word alignments, but the alignments produced by the convtransformer appear to be slightly less noisy; ($iii$) for multilingual translation of distant languages (FR+ZH$\rightarrow$EN, Figure \ref{fig:att-frzh-en}), the character alignments of the transformer become visually much noisier and concentrate on a few individual characters, with many word alignments dissolving. The convtransformer character alignments remain more spread out, and word alignment appears to be better preserved. This is another indication that the convtransformer is more robust for multilingual translation of distant languages. (iv) for multilingual translation with three inputs, where two of the three languages are close (FR+ES+ZH$\rightarrow$EN, Figure \ref{fig:att-freszh-en}), we observe a similar pattern, with word alignments being better preserved by the convtransformer.

\begin{table*}
\centering
\small
\resizebox{\textwidth}{!}{
\begin{tabular}{p{2.5cm}p{10.5cm}}
\toprule
source & Pour que ce cadre institutionnel soit efficace, il devra remédier aux lacunes en matière de réglementation et de mise en œuvre qui caractérisent à ce jour la gouvernance dans le domaine du développement durable. \\
reference & For this institutional framework to be effective, it will need to fill the regulatory and implementation deficit that has thus far characterized governance in the area of sustainable development.
 \\
\multicolumn{1}{l}{FR$\rightarrow$EN} & \\
\cmidrule{1-1}
transformer & To ensure that this institutional framework is effective, it will need to address regulatory and implementation gaps that characterize governance in sustainable development. \\
convtransformer & In order to ensure that this institutional framework is effective, it will have to address regulatory and implementation gaps that characterize governance in the area of sustainable development. \\
\midrule
\midrule
\multicolumn{1}{l}{FR+ES$\rightarrow$EN} & \\
\cmidrule{1-1}
transformer & To ensure that this institutional framework is effective, it will need to address gaps in regulatory and implementation that characterize governance in the area of sustainable development.\\
convtransformer & In order to ensure that this institutional framework is effective, it will be necessary to address regulatory and implementation gaps that characterize governance in sustainable development so far. \\
\midrule
\midrule
\multicolumn{1}{l}{FR+WB$\rightarrow$EN} & \\
\cmidrule{1-1}
transformer & To ensure that this institutional framework is effective, gaps in regulatory and implementation that have characterized governance in sustainable development to date.\\
convtransformer & For this institutional framework to be effective, it will need to address gaps in regulatory and implementation that characterize governance in the area of sustainable development. \\
\midrule
\midrule
\multicolumn{1}{l}{FR+ES+WB$\rightarrow$EN} & \\
\cmidrule{1-1}
transformer & To ensure that this institutional framework is effective, it will need to address regulatory and implementation gaps that are characterized by governance in the area of sustainable development.\\
convtransformer & If this institutional framework is to be effective, it will need to address gaps in regulatory and implementation that are characterized by governance in the area of sustainable development. \\
\bottomrule
\end{tabular}
}
\caption{Example character-level translation outputs on the UN dataset, FR$\rightarrow$EN.}
\label{tab:FR-EN}
\end{table*}
\begin{table*}
\centering
\small
\resizebox{\textwidth}{!}{
\begin{tabular}{p{2.5cm}p{10.5cm}}
\toprule
source & Estamos convencidos de que el futuro de la humanidad en condiciones de seguridad, la coexistencia pacífica, la tolerancia y la reconciliación entre las naciones se verán reforzados por el reconocimiento de los hechos del pasado. \\
reference & We strongly believe that the secure future of humanity, peaceful coexistence, tolerance and reconciliation between nations will be reinforced by the acknowledgement of the past. \\
\multicolumn{1}{l}{ES$\rightarrow$EN} & \\
\cmidrule{1-1}
transformer & We are convinced that the future of humanity in conditions of security, peaceful coexistence, tolerance and reconciliation among nations will be strengthened by recognition of the facts of the past. \\
convtransformer & We are convinced that the future of humanity under conditions of safe, peaceful coexistence, tolerance and reconciliation among nations will be reinforced by the recognition of the facts of the past. \\
\midrule
\midrule
\multicolumn{1}{l}{FR+ES$\rightarrow$EN} & \\
\cmidrule{1-1}
transformer & We are convinced that the future of mankind under security, peaceful coexistence, tolerance and reconciliation among nations will be strengthened by the recognition of the facts of the past. \\
convtransformer & We are convinced that the future of humanity in safety, peaceful coexistence, tolerance and reconciliation among nations will be reinforced by the recognition of the facts of the past. \\
\midrule
\midrule
\multicolumn{1}{l}{ES+WB$\rightarrow$EN} & \\
\cmidrule{1-1}
transformer & We are convinced that the future of humanity in safety, peaceful coexistence, tolerance and reconciliation among nations will be strengthened by the recognition of the facts of the past. \\
convtransformer & We are convinced that the future of humanity in safety, peaceful coexistence, tolerance and reconciliation among nations will be strengthened by the recognition of the facts of the past. \\
\midrule
\midrule
\multicolumn{1}{l}{FR+ES+WB$\rightarrow$EN} & \\
\cmidrule{1-1}
transformer & We are convinced that the future of mankind in safety, peaceful coexistence, tolerance and reconciliation among nations will be strengthened by the recognition of the facts of the past. \\
convtransformer & We are convinced that the future of mankind in security, peaceful coexistence, tolerance and reconciliation among nations will be strengthened by the recognition of the facts of the past. \\
\bottomrule
\end{tabular}
}
\caption{Example character-level translation outputs on the UN dataset, ES$\rightarrow$EN.}
\label{tab:ES-EN}
\end{table*}
\begin{CJK}{UTF8}{gbsn}
\begin{table*}[!htbp]
\centering
\small
\resizebox{\textwidth}{!}{
\begin{tabular}{p{2.5cm}p{10.5cm}}
\toprule
source ZH & 利用专家管理农场对于最大限度提高生产率和灌溉水使用效率也是重要的。 \\
source ZH & tjh$|$et fny$|$pe tp$|$gj pei$|$fnrt cf$|$gf jb$|$dd bv$|$ya rj$|$ym tg$|$u$|$yx t iak$|$ivc$|$ii wgkq0$|$et uqt$|$yx bn j tgj$|$s r . \\
reference EN & The use of expert farm management is also important to maximize land productivity and efficiency in the use of irrigation water. \\ 
\multicolumn{1}{l}{ZH$\rightarrow$EN} & \\
\cmidrule{1-1}
transformer & The use of expert management farms is also important for maximizing productivity and irrigation use.  \\
convtransformer & The use of experts to manage farms is also important for maximizing efficiency in productivity and irrigation water use.\\
\midrule
\midrule
\multicolumn{1}{l}{FR+ZH$\rightarrow$EN} & \\
\cmidrule{1-1}
transformer & The use of expert management farms is also important for maximizing productivity and efficiency in irrigation water use. \\
convtransformer & The use of expert management farms is also important for maximizing productivity and irrigation water efficiency. \\
\midrule
\midrule
\multicolumn{1}{l}{ES+ZH$\rightarrow$EN} & \\
\cmidrule{1-1}
transformer & The use of expert farm management is also important for maximizing productivity and irrigation water use efficiency. \\
convtransformer & The use of expert management farms to maximize efficiency in productivity and irrigation water use is also important. \\
\midrule
\midrule
\multicolumn{1}{l}{FR+ES+ZH$\rightarrow$EN} & \\
\cmidrule{1-1}
transformer & The use of expert management farms is also important for maximizing productivity and irrigation water use. \\
convtransformer & It is also important that expert management farms be used to maximize efficiency in productivity and irrigation use. \\
\bottomrule
\end{tabular}
}
\caption{Example character-level translation outputs on the UN dataset, ZH$\rightarrow$EN. }
\label{tab:CN-EN}
\end{table*}
\end{CJK}

\end{document}